\documentclass{article}

\PassOptionsToPackage{square, numbers, sort}{natbib}

\usepackage[preprint]{neurips_2019}

\usepackage[utf8]{inputenc} 
\usepackage[T1]{fontenc}    
\usepackage[hyphens]{url}   
\usepackage[hidelinks]{hyperref}       
\usepackage{booktabs}       
\usepackage{amsfonts}       
\usepackage{nicefrac}       
\usepackage{microtype}      
\usepackage{xcolor}
\usepackage{graphicx}
\usepackage{tabularx}
\usepackage{siunitx}
\usepackage{subcaption}
\usepackage{tipa}
\usepackage{tikz}
\usepackage{pgfplots}
\usepackage{multirow}
\usepackage{arydshln}
\usepackage{amssymb}
\usepackage{authblk}
\usepackage{enumitem}

\usepackage[noabbrev, capitalize]{cleveref}

\definecolor{g-blue}{HTML}{2E86C1}
\definecolor{g-red}{HTML}{B03A2E}
\definecolor{g-purple}{HTML}{AF7AC5}
\definecolor{mypink1}{rgb}{0.858, 0.188, 0.478}

\newcommand{\mypink}[1]{\textcolor{mypink1}{#1}}
\newcommand{\ignore}[1]{}

\title{TTTTTackling WinoGrande Schemas}

\author[1]{Sheng-Chieh Lin\thanks{Contributed equally.}\hspace{0.14cm}}
\newcommand\CoAuthorMark{\footnotemark[\arabic{footnote}]} 
\author[1]{Jheng-Hong Yang\protect\CoAuthorMark \hspace{0.14cm}}
\author[2]{Rodrigo Nogueira}
\author[1]{\\Ming-Feng Tsai}
\author[1]{Chuan-Ju Wang}
\author[2]{Jimmy Lin}
\affil[1]{Research Center for Information Technology Innovation, Academia Sinica}
\affil[2]{David R. Cheriton School of Computer Science, University of Waterloo}

\begin{document}

\maketitle


\begin{abstract}
We applied the T5 sequence-to-sequence model~\cite{T5} to tackle the AI2 WinoGrande Challenge~\cite{sakaguchi2019winogrande} by decomposing each example into two input text strings, each containing a hypothesis, and using the probabilities assigned to the ``entailment'' token as a score of the hypothesis.
Our first (and only) submission to the official leaderboard yielded 0.7673 AUC on March 13, 2020, which is the best known result at this time and beats the previous state of the art by over five points.\footnote{\url{https://leaderboard.allenai.org/winogrande/submissions/public}}\end{abstract}

\section{Introduction}

Other than encoder-only pretrained transformer architectures~\cite{Liu:1907.11692:2019, BERT, XLNET}, encoder--decoder style pretrained transformers~\cite{T5, lewis2019bart} have been proven to be effective in text generation tasks as well as comprehension tasks.
This paper describes our submission to the commonsense reasoning task leaderboard of the AI2 WinoGrande Challenge~\cite{sakaguchi2019winogrande}, which uses the text-to-text transfer transformer (T5); our approach currently represents the state of the art.

In T5~\cite{T5}, NLP tasks are formulated as text-to-text problems, where the inputs are cast into natural language templates that contain the task descriptors.
Concretely, Raffel et al. provide the following example for MNLI~\cite{williams-etal-2018-broad}, where the goal is to predict whether a premise implies (``entailment'') or contradicts (``contradiction'') a hypothesis, or neither (``neutral'').
Thus, a training example becomes:

\begin{quote}``mnli premise: I hate pigeons. hypothesis: My feelings towards pigeons are filled with animosity.''
\end{quote}

with ``entailment'' as the corresponding ground truth target output.
In other words, a token representing each class is directly used as the prediction target.

\section{Approach}

The natural language template approach enables various options to formulate the WinoGrande commonsense reasoning task as a text-to-text problem with T5.
Here we adopt a formulation similar to the MNLI template.
Consider a concrete example:

\begin{quote}
He never comes to my home, but I always go to his house because the \_ is smaller. \\
Option1: home; Option2: house
\end{quote}

\noindent In this case, the correct replacement for \_ is Option1.
We decompose the above problem into two source--target training examples, where \_ is replaced with each option and annotated with the correct answer as the target token, as shown in Table~\ref{tb:training-example}.
In addition, we reformulate each example into a commonsense reasoning ``template'' with two statements:\ hypothesis (from \_ to the end of the original problem statement) and premise (the remaining part of the original problem statement). 
Note that the bold and colored fonts are for clarity only; those tokens are not marked in any way in the model input.

\begin{table*}[htb!]
\centering
\begin{tabular}{p{0.6\textwidth}p{0.2\textwidth}}
\toprule
Source & Target \\
\midrule
\textbf{hypothesis:} \mypink{\emph{home}} is smaller. \textbf{premise: }He never comes to my home, but I always go to his house because the & \textbf{entailment} \\
\textbf{hypothesis:} \mypink{\emph{house}} is smaller. \textbf{premise:} He never comes to my home, but I always go to his house because the & \textbf{contradiction} \\
\bottomrule
\end{tabular}
\caption{Decomposing WinoGrande problems into training instances for T5.
\vspace{0.25cm}}
\label{tb:training-example}
\end{table*}

At inference (test) time, we also decompose the problem into two inputs, where each input is formulated in exactly the same manner as in Table~\ref{tb:training-example}, with either one of the answer options.
We then feed each into T5 to predict a target token.
In this scenario, there are four possible outcomes:

\begin{enumerate}[leftmargin=*]

\item one produces ``entailment'' and the other ``contradiction'',

\item one produces ``entailment'' or ``contradiction'' and the other some other token,

\item both produce some other tokens, and

\item both produce the same token, either ``entailment'' or ``contradiction''.
    
\end{enumerate}

Ideally, T5 would produce contrastive tokens for each input pair, as in case (1), which allows us to unambiguously select the final answer. 
However, the model might produce the same tokens for each input, or even tokens not in the predefined set, as in cases (2) to (4).
To deal with these cases, we apply a softmax over the logits of the pair of predefined target tokens, similar to Nogueira et al.~\cite{Nogueira_etal_arXiv2020_T5}.
From this, we can compute the probabilities of the predefined target tokens (in the case of Table~\ref{tb:training-example}, ``entailment'' and ``contradiction'').
Then, we compare the probabilities across both input instances, and in cases (2) to (4), we select the instance that has a higher probability as the correct answer.

This general problem setup allows us to choose the target tokens, which may have an impact on the prediction accuracy~\cite{Nogueira_etal_arXiv2020_T5}.
In addition to selecting ``entailment'' vs.\ ``contradiction'' as the target, we also tried the contrastive pair ``true'' vs.\ ''false''.

In our experiment, we fine-tune T5-3B on Google Colab's TPU v2 with a batch size of 16, a learning rate of $2 \cdot 10^{-4}$, and save model checkpoints every 5000 steps. It takes 130k steps to converge for the XL data size (see below).
At inference time, we use greedy decoding and select for evaluation the model checkpoint that achieves the highest score on the development set.
We did not experiment with T5-11B due to limited computational resources.

\section{Results}

Experimental results on the WinoGrande development set are reported in Table~\ref{tab:main_results} for different training data sizes.
Note that we fine-tune the model for each training data size separately.
A \checkmark under the ``logit'' column indicates that we used the softmax over the target tokens as described above.
Without this technique, given the original two-choice question, if T5 outputs the same tokens for the two processed inputs, we simply assign Option1 as the answer.
The table also reports ``zero-shot'' performance, i.e., performing inference on the development set without any model fine tuning.
Condition \#2 represents our submission to the official leaderboard, which achieves 0.7673 AUC on the held-out test set.

From these results, we see that the logit trick clearly improves performance, which is consistent with the observations of Nogueira et al.~\cite{Nogueira_etal_arXiv2020_T5}.
In fact, applying this technique in the zero-shot setting yields performance that is clearly better than random.
Another interesting finding is that the choice of target token appears to have an impact on performance, which is also consistent with the above work.
Since using true/false as the target token (conditions \#3 and \#4) did not improve performance much over conditions with entailment/contradiction, we did not run all data size conditions given our limited computational resources.

\begin{table*}[t]
\centering
\resizebox{\textwidth}{!}{
\begin{tabular}{llccccccc}
\toprule
&\multicolumn{2}{c}{Condition} & \multicolumn{6}{c}{Training size} \\
\cmidrule(lr){2-3}\cmidrule(lr){4-9}
Condition &Answer token & logit & Zero-Shot & XS & S & M & L & XL\\
\toprule
\#1&\multirow{2}{*} {entailment/contradiction} & & 0.506 & 0.657 & 0.693 & 0.757 & 0.809 & 0.840\\
\#2&& \checkmark & 0.608 & 0.718 & 0.740 & 0.788 & 0.837 & 0.854\\
  \midrule
\#3&\multirow{2}{*} {true/false} & & 0.477 & 0.676 & - & - & - & 0.852\\
\#4& & \checkmark & 0.566 & 0.723 & - & - & - & 0.865\\
\midrule
\multicolumn{3}{l}{Our leaderboard results on test set} & - & 0.683 & 0.705 & 0.776 & 0.824 & 0.846 \\
\bottomrule
\end{tabular}
}
\caption{Main results on the WinoGrande development set. Condition \#2 is our current submission.\vspace{0.25cm}}
\label{tab:main_results}
\end{table*}

Looking at the current WinoGrande leaderboard, it appears that the previous state of the art is based on RoBERTa~\cite{Liu:1907.11692:2019}, which can be characterized as an encoder-only transformer architecture.
Since T5-3B is larger than RoBERTa, it cannot be ruled out that model size alone explains the performance gain.
However, when coupled with the observations of Nogueira et al.~\cite{Nogueira_etal_arXiv2020_T5}, T5's ``generative capability'', i.e., its ability to generate fluent text, honed through pretraining, seems to play an important role.
The fact that the choice of target tokens affects prediction accuracy is consistent with this observation.
How and why is the subject of ongoing work.

\section{Implications}

Collectively, the success of large pretrained neural models, both encoder-only BERT-like architectures as well as encoder--decoder architectures like T5, raise interesting questions for the pursuit of commonsense reasoning.
Researchers have discovered that previous models perform well on benchmark datasets because they pick up on incidental biases in the dataset that have nothing to do with the task; in contrast, the WinoGrande dataset has devoted considerable effort to reducing such biases, which may allow models to (inadvertently) ``cheat'' (for example, using simple statistical associations).
While it is certainly true that datasets over-estimate the commonsense reasoning capabilities of modern models~\cite{sakaguchi2019winogrande}, there are alternative and complementary explanations as well:

It has been a fundamental assumption of the research community that commonsense reasoning is difficult because it comprises tacit rather than explicit knowledge~\cite{WINOGRAD19721}.
That is, commonsense knowledge---like water is wet and that a tuba is usually too big to fit in a backpack---is not written down anywhere (unlike, say, factual knowledge, which can be modeled in a knowledge graph).
As a result---the reasoning goes---data-driven techniques (even neural models) will be of limited use due to the paucity of {\it relevant} corpora.

Yet, previous encoder-only architectures like RoBERTa that exploit a language modeling objective (that is, relying {\it only} on explicit textual knowledge) can clearly make headway in a commonsense reasoning task, and we can further improve upon these approaches with a sequence-to-sequence model.
This leaves us with two possible explanations:\ despite careful controls, the WinoGrande challenge {\it still} contains incidental biases that these more sophisticated models can exploit, or that we are genuinely making at least {\it some progress} in commonsense reasoning.
The latter, in particular, challenges the notion that commonsense knowledge is (mostly) tacit.
Perhaps it is the case that in a humongous corpus of natural language text, someone really has written about trying to stuff a tuba in a backpack?

\section*{Acknowledgments}

This research was supported in part by the Canada First Research Excellence Fund and the Natural Sciences and Engineering Research Council (NSERC) of Canada.
We would like to thank Google Colab for providing support in terms of computational resources.

\bibliographystyle{abbrv}
\bibliography{T5sense}

\end{document}